\documentclass[letterpaper, 10 pt, conference]{ieeeconf}
\IEEEoverridecommandlockouts
\usepackage{multirow}
\usepackage{cite}
\usepackage{amsmath,amssymb,amsfonts}
\usepackage{algorithmic}
\usepackage{graphicx}
\usepackage{textcomp}
\usepackage{xcolor}
\usepackage{comment}
\def\BibTeX{{\rm B\kern-.05em{\sc i\kern-.025em b}\kern-.08em
    T\kern-.1667em\lower.7ex\hbox{E}\kern-.125emX}}
\begin{document}

\title{Levelling the Playing Field: A Comprehensive Comparison of Visual Place Recognition Approaches under Changing Conditions}

\author{Mubariz Zaffar$^{1}$, Ahmad Khaliq$^{1}$, Shoaib Ehsan$^{1}$,  Michael Milford$^{2}$ and Klaus McDonald-Maier$^{1}$
\thanks{This work is supported by the UK Engineering and Physical Sciences Research Council through grants EP/R02572X/1 and EP/P017487/1.}
	\thanks{$^{1}$Authors are with the Embedded and Intelligent Systems Laboratory in Computer Science and Electronic Engineering department,
		University of Essex, Colchester, United Kingdom.
		{\tt\small \{mubariz.zaffar,ahmad.khaliq,sehsan,kdm\} @essex.ac.uk}}%
	\thanks{$^{2}$Michael Milford is with the Australian Centre for Robotic Vision and School of Electrical Engineering and Computer Science, Queensland University of Technology, Brisbane, Australia.
		{\tt\small michael.milford@qut.edu.au}}
}

\maketitle

\begin{abstract}
In recent years there has been significant improvement in the capability of Visual Place Recognition (VPR) methods, building on the success of both hand-crafted and learnt visual features, temporal filtering and usage of semantic scene information. The wide range of approaches and the relatively recent growth in interest in the field has meant that a wide range of datasets and assessment methodologies have been proposed, often with a focus only on precision-recall type metrics, making comparison difficult. In this paper we present a comprehensive approach to evaluating the performance of $10$ state-of-the-art recently-developed VPR techniques, which utilizes three standardized metrics: (a) Matching Performance b) Matching Time c) Memory Footprint. Together this analysis provides an up-to-date and widely encompassing snapshot of the various strengths and weaknesses of contemporary approaches to the VPR problem. The aim of this work is to help move this particular research field towards a more mature and unified approach to the problem, enabling better comparison and hence more progress to be made in future research.
\end{abstract}

\begin{keywords}
Visual Place Recognition, comparison, state-of-the-art, Seq-SLAM, VLAD, HybridNet, BoW
\end{keywords}

\section{Introduction}
Simultaneous Localization and Mapping (SLAM) represents the ability of a robot to create a map of its environment and concurrently localize itself within it \cite{cadena2016past}. In a monocular SLAM system \cite{eade2006scalable}, the only source of information is a camera and thereby places in the environment are represented as images. Thus for loop closure, a robot needs to be able to successfully match images of the same place upon repeated traversals. Such an ability of the robot to remember a previously visited place to perform loop-closure is termed and researched as Visual Place Recognition (VPR). 

\begin{figure}[t]
\begin{center}
\includegraphics[width=1.0\linewidth, height=0.5\linewidth]{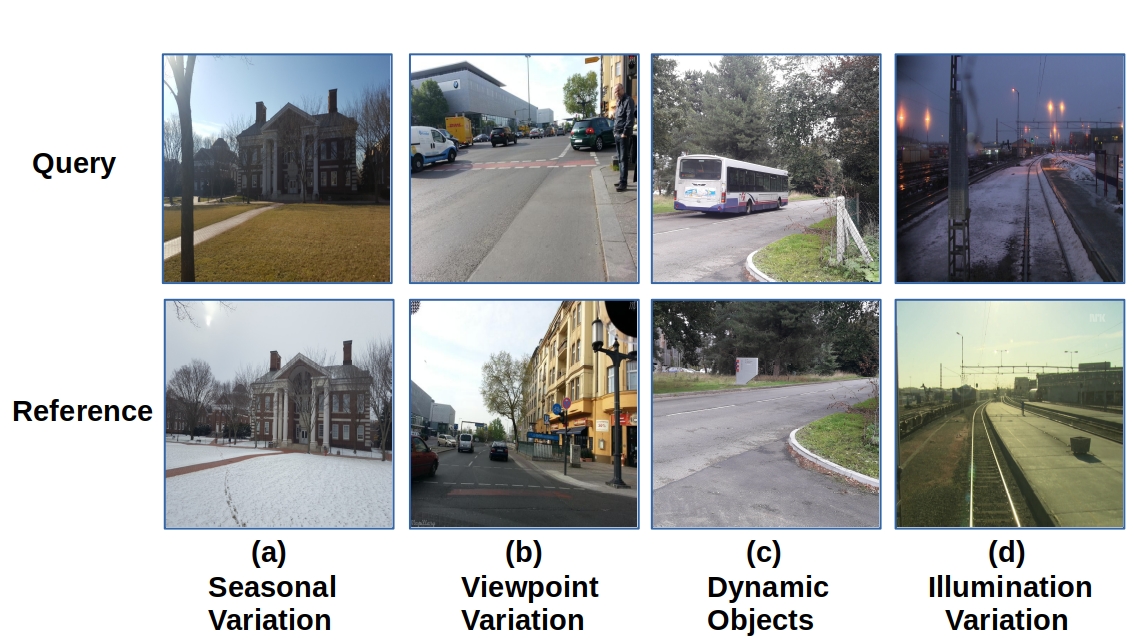}
\end{center}
\caption{Variations in the appearance of places are illustrated where; (a) Seasonal variation observed from summer to winter in the same place (b) Change in camera viewpoint leading to drastic change in observed structures (c) Commonly seen dynamic objects in urban scenes (d) Appearance change as a result of day-to-night transition.}
\label{Fig:VariationsInPlaces}
\end{figure}

VPR is a well-understood problem and acts as an important module of a Visual-SLAM based autonomous system \cite{vprasurvey}. However, VPR is highly challenging due to the significant variations in appearance of places under changing conditions. Throughout VPR research over the past years, we see $4$ such variations in appearances of places, which have been widely discussed and tackled by different novel VPR techniques. \textbf{Seasonal Variation:} Appearance of places change drastically from summer to winter or spring to autumn posing challenges for state-of-the-art VPR techniques \cite{naseer2014robust}\cite{valgren2010sift}. \textbf{Viewpoint Variation:} Images of the same place may look very different when taken from different viewpoints \cite{pronobis2006discriminative}. This viewpoint variation could be a simple lateral variation or a more complex angular variation coupled with changes in focus, base point and/or zoom during repeated traversals \cite{garg2018lost}. \textbf{Illumination Variation:} This is the result of daylight changes and intermediary transitions during different times of the day/night, which make place recognition difficult to perform \cite{ranganathan2013towards}\cite{milford2015sequence}. \textbf{Dynamic Objects:} Objects such as cars, people, animals etc. can also change the appearance of a scene and thus a VPR technique should be able to suppress any features coming from these dynamic objects \cite{wang2007simultaneous}\cite{naseer2017semantics}. We show all these variations in Fig. \ref{Fig:VariationsInPlaces}.   

While many different techniques \cite{sunderhauf2015performance}\cite{sunderhauf2015place} \cite{panphattarasap2016visual} have been proposed to tackle each (or a combination) of the $4$ variations, a thorough and holistic comparison of these techniques is needed for an up-to-date review. In this paper, we take up the task to evaluate $10$ novel VPR techniques on the most challenging public datasets using the same platform, evaluation metric and ground truth data. While presenting the matching performance and (more recently) matching time has been common in VPR research; we additionally enlist the memory footprint of these VPR techniques which is an essential factor at deployment. The novel contributions of this paper are as follows:

\begin{enumerate}
    \item The techniques compared in this paper encompass years of VPR research and a comparison of such magnitude has not been reported previously.
    \item VPR performance is highly sensitive to the choice of evaluation datasets, computational platform, evaluation metric and ground truth data. By keeping all of these variables constant, we bring VPR techniques to a common ground for evaluation.
    \item Memory footprint for map creation at deployment time is a critical factor and thus, we report the feature vector size for all $10$ VPR techniques.  
    
\end{enumerate}

The rest of the paper is organized as follows. Section II provides a detailed literature review of the VPR techniques both from handcrafted and CNN-based descriptor paradigms. In section III, we  describe the experimental setup and parametric configurations employed for analyzing the performance of contemporary VPR techniques. Section IV presents the detailed analysis and results obtained by evaluating the targeted frameworks on challenging benchmark datasets. Finally, conclusions are presented is Section V. 

\section{Literature Review}
Visual Place Recognition (VPR) has seen significant advances at the frontiers of matching performance and computational superiority over the past few years. While the former has been the prime focus of all VPR techniques, latter has been discussed only recently. A detailed survey of the challenges, developments and future directions in VPR has been performed by Lowry et al. \cite{vprasurvey}.

The early techniques employed for image matching in VPR consisted of handcrafted feature descriptors \cite{bay2006surf}\cite{lowe2004distinctive}. These handcrafted descriptors could then be classified into either local or global feature descriptors. SIFT (Scale Invariant Feature Transform \cite{lowe2004distinctive}) is a local feature descriptor that extracts keypoints from an image using difference-of-gaussians and describes these keypoints using histogram-of-oriented-gradients. It has been used for VPR by Stumm et al. in \cite{stumm2013probabilistic}. SURF (Speeded Up Robust Features) which is a modified version of SIFT uses Hessian-based detectors instead of Harris detectors and was introduced by Bay et al. in \cite{bay2006surf}. SURF is used for VPR by authors in \cite{murillo2007surf}. Other handcrafted techniques used in VPR include Centre Surrounded Extremas (CenSurE \cite{agrawal2008censure}) which uses centre surrounded filters across multiple scales at each image pixel for keypoint search, FAST \cite{rosten2006machine} which is a corner detector utilizing a corner response function (CRF) to find the best corner candidates, and Bag of Visual Words (BoW \cite{sivic2003video}) which creates a visual vocabulary of image patches (or patch descriptors) . Gist \cite{oliva2006building} is a global feature detector employing Gabor filters and has been used for image matching by authors in \cite{murillo2009experiments} and \cite{singh2010visual}. WI-SURF is a global variant of SURF and has been used for real-time visual localization in \cite{badino2012real}. Histogram-of-oriented-gradients (HOG) \cite{MERL_TR9403} \cite{dalal2005histograms} computes gradients at all image pixels and constructs a histogram with bins classified by gradient angles and containing sums of gradient magnitudes. HOG is used for VPR by McManus et al. in \cite{mcmanus2014scene}.  Seq-SLAM \cite{seqslam} is a VPR technique using confusion matrix created by subtracting patch-normalized sequences of frames to find the best matched route.

Similar to many other applications \cite{tolias2015particular}\cite{liu2017cross}, the use of neural networks in VPR achieved far better results than any handcrafted feature descriptor based approach. This was studied by Chen et al. in \cite{chen2014convolutional}, where given an input image to a pre-trained convolutional neural network (CNN), they extracted features from layers' responses and subsequently used these features for image comparison \cite{sermanet2013overfeat}. Following-up on their previous work, they trained two dedicated CNNs (namely AMOSNet and HybridNet) in \cite{chen2017deep} on Specific Places Dataset (SPED) achieving state-of-the-art VPR performance. Both AMOSNet and HybridNet have the same architecture as CaffeNet \cite{krizhevskyimagenet}, where the weights of former were randomly initialized while the latter used weights from CaffeNet trained on ImageNet dataset \cite{deng2009imagenet}. Description of features/activations of convolutional layers has also been  researched, with the advent of pooling approaches employed on convolution layers including Max- \cite{tolias2015particular}, Sum- \cite{babenko2015aggregating}, Spatial Max-Pooling \cite{jaderberg2015spatial} and Cross-Pooling \cite{liu2017cross}. While CNN based features proved highly invariant to environmental changes, the CNN architecture is designed for classification purpose and does not output a feature descriptor given an input image. Thus, Arandjelovic et al. \cite{arandjelovic2016netvlad} added a new VLAD (Vectors of Locally Aggregated Descriptors) layer to the CNN architecture which could be trained in an end-to-end manner for VPR. This new VLAD layer was then amended to different CNN models (including AlexNet and VGG-16) and trained on place-recognition dataset captured from Google Street View Time Machine. The unavailability of large labelled datasets for training VPR specific neural networks limits the environmental variations seen by such a CNN; thus \cite{merrill2018lightweight} proposed a new unsupervised VPR-specific training mechanism utilizing a convolutional auto-encoder and HOG descriptors. For images containing repetitive structures, Torii et al. \cite{torii2013visual} proposed a robust mechanism for collecting visual words into descriptors. Synthetically created views are utilized for illumination invariant VPR in \cite{torii201524}, which shows that highly conditionally variant images can still be matched if they are from the same viewpoint.

More recently \cite{chen2018learning}\cite{khaliqholistic}\cite{facil2019condition}\cite{hausler2019multi}, different VPR research works have suggested that regions based description of features can substantially increase matching performance by focusing on salient regions and discarding confusing regions. The work in \cite{tolias2015particular} has revisited the CNN based descriptors efficient for image search, succeeded by geometric re-ranking and query expansion. In particular, several image regions are encoded employing Max-Pooling over the convolutional layers' responses, named as regional maximum activation of convolutions (R-MAC). Similar to Cross-Pooling \cite{liu2017cross}, authors in \cite{chen2017only} employed a cross-convolution technique for VPR to pool features from the convolutional layers.They first find salient region proposals from late convolutional layers of object centric VGG-16 \cite{simonyan2014very} and select top $200$ energetic regions. These regions are then described by using activations from prior convolutional layers. Furthermore, a separate $5k$ dataset is employed to learn a $10k$ regional dictionary for BoW \cite{sivic2003video} features encoding scheme, thus, named as Cross-Region-BoW. Despite the recent state-of-the-art (SOTA) performances of deep CNNs for VPR and other image retrieval tasks, the significant computation- and memory-overhead limits their practical deployment. Khaliq et al. \cite{khaliq2018holistic} proposed a lightweight CNN-based regional approach combined with VLAD (using a separate visual word vocabulary learned from $2.6k$ dataset), Region-VLAD has shown boost-up in image retrieval speed and accuracy.

\section{Experimental Setup}
This section first discusses the contemporary VPR techniques that are compared in this paper. We then present the datasets used for evaluation. Finally, we describe the evaluation metrics considered for comparison in our work.    

\subsection{VPR Techniques} \label{vpr_techniques}

\subsubsection{HOG Descriptor}
Histogram-of-oriented-gradients (HOG) is one of the most widely used handcrafted feature descriptor. While it does not perform nearly well to any other VPR technique, it is a good starting point for a comparison such as ours. Our motivation to select HOG is also based upon its performance as shown by McManus et al.  \cite{mcmanus2014scene} and the utility it offers an an underlying feature descriptor for training a convolutional auto-encoder in \cite{merrill2018lightweight}. We use a cell size of $8 \times 8$ and a block size of $16 \times 16$ with total number of histogram bins equal to $9$. HOG descriptors of two images are subsequently compared using cosine similarity.

\subsubsection{Seq-SLAM}
Seq-SLAM showed excellent immunity to seasonal and illumination variations by using sequential information to its advantage. The proposed implementation has been open-sourced in MATLAB and ported to Python. We use a sequence size of $10$, minimum velocity of $0.8$ and max velocity of $1.2$ for evaluating Seq-SLAM.   

\subsubsection{AlexNet}
S{\"u}nderhauf et al. studied the performance of features extracted from AlexNet and found \textit{conv3} to be the most robust to environmental variations. The activation maps are encoded into feature descriptors by using Gaussian random projections. Our implementation of AlexNet is similar to the one presented by authors in \cite{merrill2018lightweight}.    

\subsubsection{NetVLAD}
We have employed the Python implementation of NetVLAD open-sourced in \cite{cieslewski2018data}. The model selected for evaluation is VGG-16 which has been trained in an end-to-end manner on Pittsburgh 250K dataset \cite{arandjelovic2016netvlad} with a dictionary size of $64$ while performing whitening on the final descriptors.      

\subsubsection{AMOSNet}
AMOSNet has been trained from scratch on SPED dataset and the model weights have been open-sourced by authors in \cite{chen2017deep}. We therefore implement spatial pyramidal pooling on pre-trained AMOSNet and use activations from \textit{conv5} to extract and describe features. L1-difference is subsequently used to match features descriptors of two images.    

\subsubsection{HybridNet}
Similar to AMOSNet, model parameters for HybridNet trained on SPED dataset have also been open-sourced. However, the weights of top-5 HybridNet convolutional layers are initialized from CaffeNet trained on ImageNet dataset. We employ spatial pyramidal pooling on activations from \textit{conv5} layer of HybridNet. Feature descriptors of two images are then matched using L1-difference.

\subsubsection{Cross-Region-BOW}
We have employed the \cite{chen2017onlyCode} open-source MATLAB implementation for experimentation; VGG-16 \cite{simonyan2014very} pre-trained on ImageNet dataset is used while employing \textit{conv5\_3} and \textit{conv5\_2} for identification and extraction of regions respectively. Image comparison is carried out by finding the best mutually matched regions and describing these regions using a $10k$ BoW dictionary.

\subsubsection{R-MAC}
The MATLAB implementation for R-MAC is available at \cite{tolias2015particularRMACCode}. We used \textit{conv5\_2} of object-centric VGG-16 for regions-based features. For a fair comparison, we remove the geometric verification block while performing power and l2 normalization on the retrieved R-MAC representations. The retrieved R-MACs are mutually matched, followed by aggregation of the mutual regions' cross-matching scores.  

\subsubsection{Region-VLAD}
We employed \textit{conv4} of HybridNet for evaluating the Region-VLAD VPR approach. The employed dictionary contains $256$ visual words used for VLAD retrieval. Cosine similarity is subsequently used for descriptor comparison. 

\subsubsection{CALC}
Merrill et al. trained a convolutional auto-encoder for the first-time in an unsupervised manner for VPR, where the objective of auto-encoder was to re-create the HOG descriptor of original image given a distorted version of the original image as input. Authors have open-sourced their implementation and we use model parameters from $100,000$ training iteration for comparison in our work.

\subsection{Evaluation Datasets}
Several different datasets  \cite{seqslam} \cite{merrill2018lightweight} \cite{chen2017only} \cite {sunderhauf2015place} \cite{3} \cite{4} \cite{5} \cite{nordlanddataset} \cite{8} \cite{10} have been proposed for evaluating VPR techniques over the years. These datasets comprise of different types of variations ranging from viewpoint, seasonal and illumination variations to a combination of these. In order to challenge and put all the VPR techniques presented in sub-section \ref{vpr_techniques} to their limits, we select $3$ datasets with the most extreme variations. This sub-section is dedicated to introducing these $3$ datasets. We also summarize the qualitative and quantitative nature of datasets in Table \ref{table:datasets}.

\renewcommand{\arraystretch}{1.2}
\renewcommand{\tabcolsep}{2.2pt}

\begin{table}[] 
\caption{BENCHMARK PLACE RECOGNITION DATASETS}
\label{table:datasets}
\begin{tabular}{|c|c|c|c|c|c|}
\hline
\multirow{2}{*}{\textbf{Dataset}} & \multicolumn{2}{c|}{\textbf{Traverse}} & \multirow{2}{*}{\textbf{Environment}}                       & \multicolumn{2}{c|}{\textbf{Variation}} \\ \cline{2-3} \cline{5-6} 
                                  & \textbf{Test}    & \textbf{Reference}   &                                                             & \textbf{Viewpoint} & \textbf{Condition} \\ \hline
Nordland                          & 172              & 172                  & Train journey                                               & strong             & very strong        \\ \hline
Berlin Kudamm                     & 222              & 201                  & Urban                                                       & very strong        & strong             \\ \hline
Gardens Point                     & 200              & 200                  & \begin{tabular}[c]{@{}c@{}}University\\ campus\end{tabular} & strong             & very strong        \\ \hline
\end{tabular}
\end{table}

\subsubsection{Berlin Kudamm Dataset}
\begin{figure}[t]
\begin{center}
\includegraphics[width=1.0\linewidth,height=0.4\linewidth]{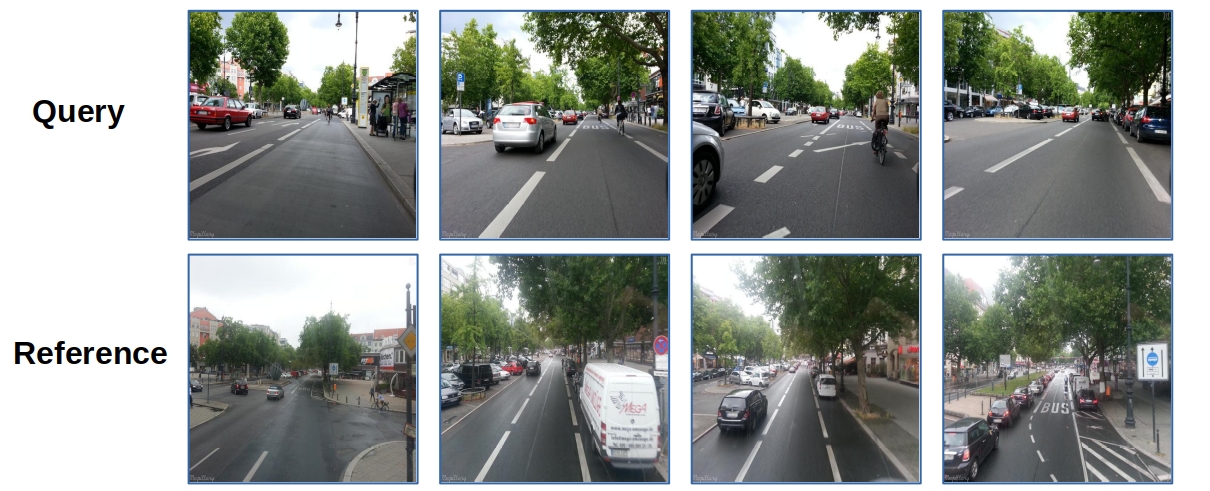}
\end{center}
\caption{Berlin Kudamm dataset sample images are shown here. The query and reference traverses exhibit extreme viewpoint variation. This dataset contains recurring and upfront dynamic objects which is uncommon to any other VPR dataset.}
\label{Fig:berlin_kudamm_samples}
\end{figure}

This dataset was introduced in \cite{chen2017only} and has been captured from crowd-sourced photo-mapping platform called \textit{Mapillary\footnote{https://www.mapillary.com/}}. Both the traverses exhibit strong viewpoint and conditional changes as visualized in Fig. \ref{Fig:berlin_kudamm_samples}.  Due to its urban nature, dynamic objects such as vehicles and pedestrians are observed in most of the captured frames. Ground truth is obtained using GPS information to build place-level correspondence. A retrieved image against a query is considered as a correct match if it is either of the $5$ closest frames in ground-truth. Thus, for a query image $q$ and its ground-truth image $n$ in the reference database, images $n-2$ to $n+2$ also serve as corresponding correct matches. 

\subsubsection{Gardens Point Dataset}
\begin{figure}[t]
\begin{center}
\includegraphics[width=1.0\linewidth,height=0.4\linewidth]{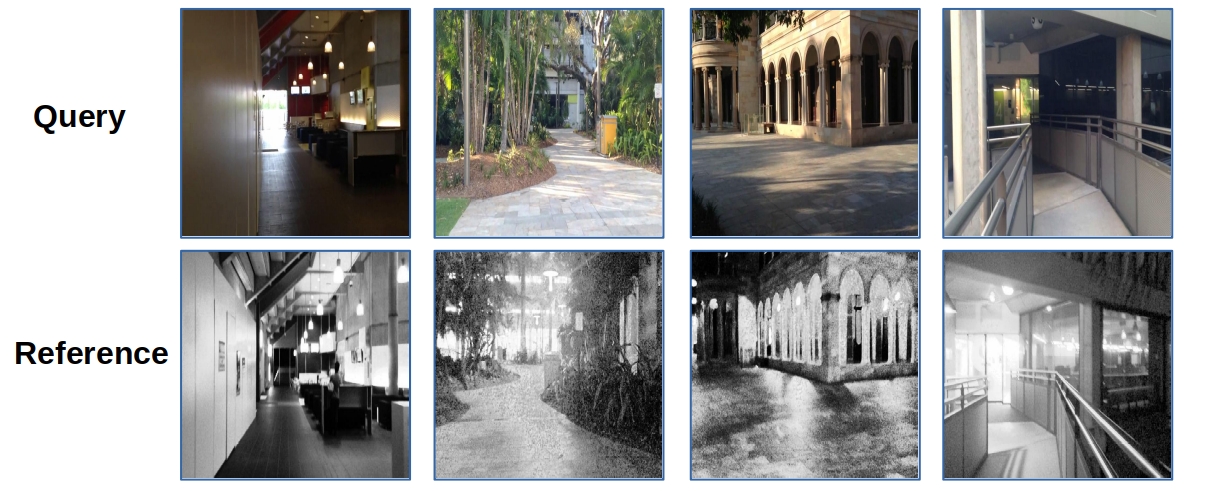}
\end{center}
\caption{Gardens Point dataset sample images are presented here. The query and reference traverses are highly illumination variant and accompanied with lateral viewpoint variation.}
\label{Fig:gardenspoint_samples}
\end{figure}

This dataset is constructed from Queensland University of Technology (QUT) with the first traverse captured during the day and the reference traverse taken at night with laterally changed viewpoint. Variations in the dataset are shown in Fig. \ref{Fig:gardenspoint_samples}. The ground truth is obtained by frame- and place-level correspondence. A retrieved image against a query is considered as a correct match if it is either of the $5$ closest frames in ground-truth. That is, for a query image $q$ and its ground-truth image $n$ in the reference database, images $n-2$ to $n+2$ also serve as corresponding correct matches.  

\subsubsection{Nordland Dataset}
\begin{figure}[t]
\begin{center}
\includegraphics[width=1.0\linewidth,height=0.4\linewidth]{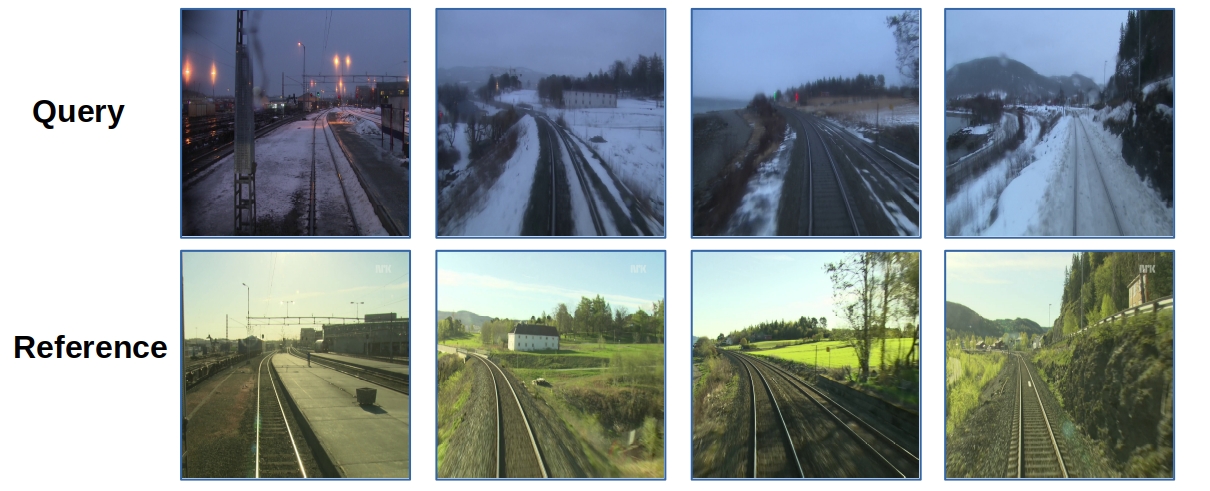}
\end{center}
\caption{Nordland dataset sample images are presented in this figure. This dataset is one of the highly seasonally variant dataset and has manually introduced lateral viewpoint variation.}
\label{Fig:nordland_samples}
\end{figure}

A train journey is captured in this dataset with the first traverse taken during winter and the second traverse during summer. While this dataset contains strong seasonal changes as shown in Fig. \ref{Fig:nordland_samples}, we introduce lateral viewpoint variation by manually cropping images. The ground truth consists of frame-level correspondence with a retrieved image against a query considered as a correct match if it is either of the $3$ closest frames in ground-truth. Thus, for a query image $q$ and its ground-truth image $n$ in the reference database, images $n-1$ to $n+1$ also serve as corresponding correct matches.

\subsection{Evaluation Metrics}

\subsubsection{Matching Performance}
In image-retrieval for VPR, area under the precision-recall curves (AUC) is a well-established evaluation metric. Although, AUC has been used widely for reporting VPR performance in literature, the computational methodology used for area computation can result in different values of AUC. We compute the precision and recall values for every matched/unmatched query image. To maintain consistency in our work, we only compute and report AUC performance by utilizing equation \ref{AUC_equation}.   

\begin{equation} \label{AUC_equation}
AUC = \sum_{i=1}^{N-1} \frac {(p_i + p_{i+1})} {2} \times (r_{i+1} - r_i)
\end{equation}

\begin{equation*}
\begin{aligned}
where; \; \; N = No. \; of \; Query \; Images
 \\
p_i=Precision \; at \; point \; i \\
r_i= Recall \; at \; point \; i 
\end{aligned}
\end{equation*}

\subsubsection{Matching Time}
For real-time autonomous robotics, matching time is an important factor to be considered at deployment. For all $10$ VPR techniques, we report the matching time of a query image given pre-computed feature descriptors of reference images. This matching time (reported in seconds) includes the feature encoding time for an input query image and the descriptor matching time for $R$ number of reference images.

\subsubsection{Memory Footprint}
The deployment use of VPR is coupled with map creation in SLAM. Therefore, the size of reference image descriptors is an important factor to be considered for the practicality of a VPR technique. While this has not been previously discussed, we enlist the size in bytes of a reference image feature descriptor for all $10$ VPR techniques. This gives a good idea about the scalability of a technique for large-scale visual place recognition.

\section{Results and Analysis}
This section is dedicated to the performance evaluation of all VPR techniques. We present the image matching performance on benchmark VPR datasets, followed by matching time and memory footprint, each discussed in their respective subsections. All evaluations are performed with an Intel(R) Xeon(R) Gold 6134 CPU @ 3.20GHz with 64GB physical memory running a Ubuntu 16.04.6 LTS. 

\subsection{Matching Performance}
\begin{figure*}[t]
\begin{center}
\includegraphics[width=1.0\linewidth]{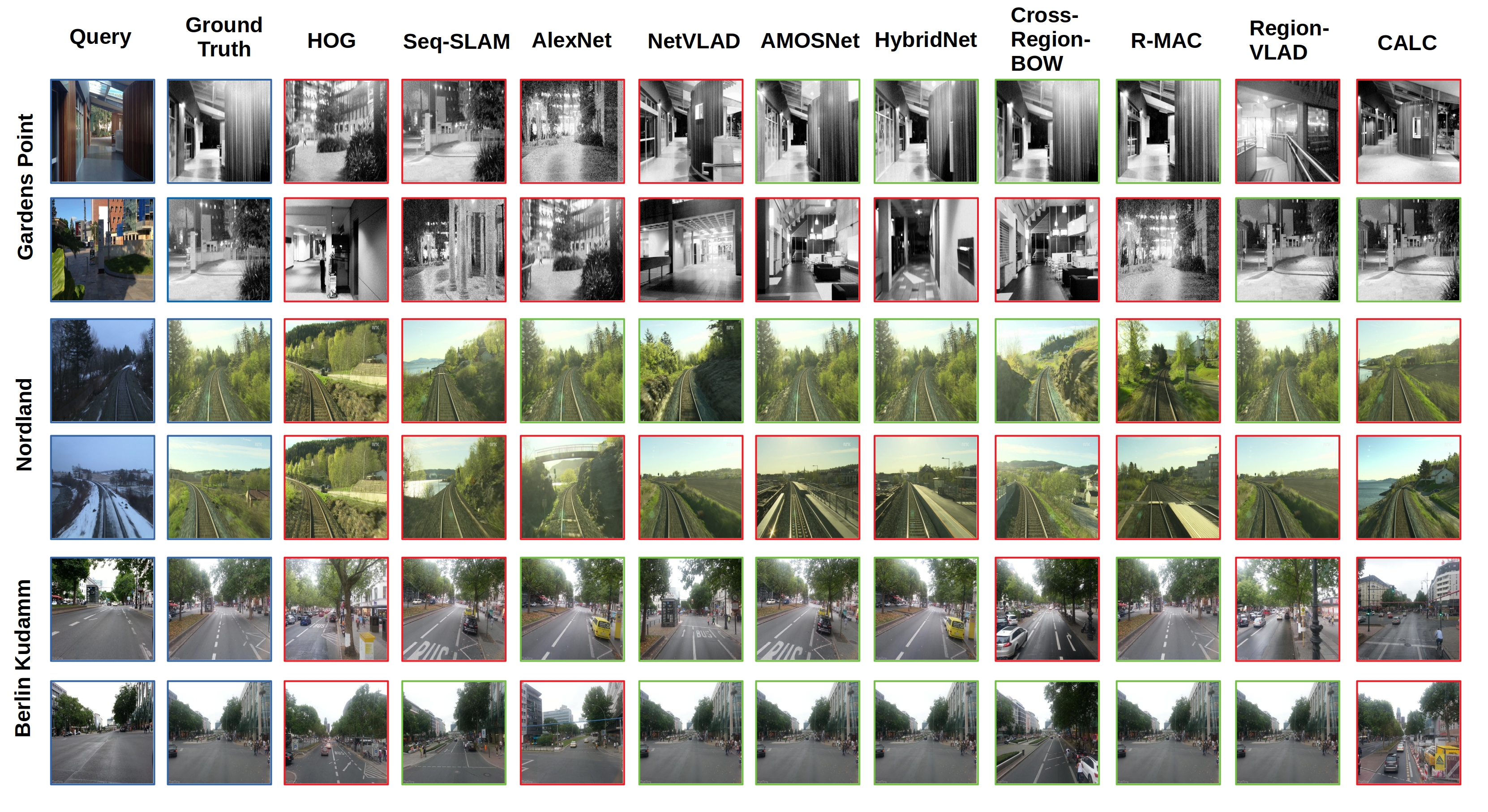}
\end{center}
\caption{Samples of images matched/mismatched by different VPR techniques on all three datasets are presented. The first two columns show the query and ground-truth reference images respectively, followed by images retrieved by each of the $10$ VPR techniques.}
\label{Fig:exemplarimages}
\end{figure*}

This sub-section reports the AUC performance of all $10$ VPR techniques on each of the $3$ datasets. We also show exemplar image matches from all three datasets in Fig. \ref{Fig:exemplarimages}. While some exemplar images have been matched by most state-of-the-art VPR techniques, we also include examples that are mismatched across the board.

\subsubsection{Berlin Kudamm Dataset}
\begin{figure*}[t]
\begin{center}
\includegraphics[width=1.0\linewidth, height=0.6\linewidth]{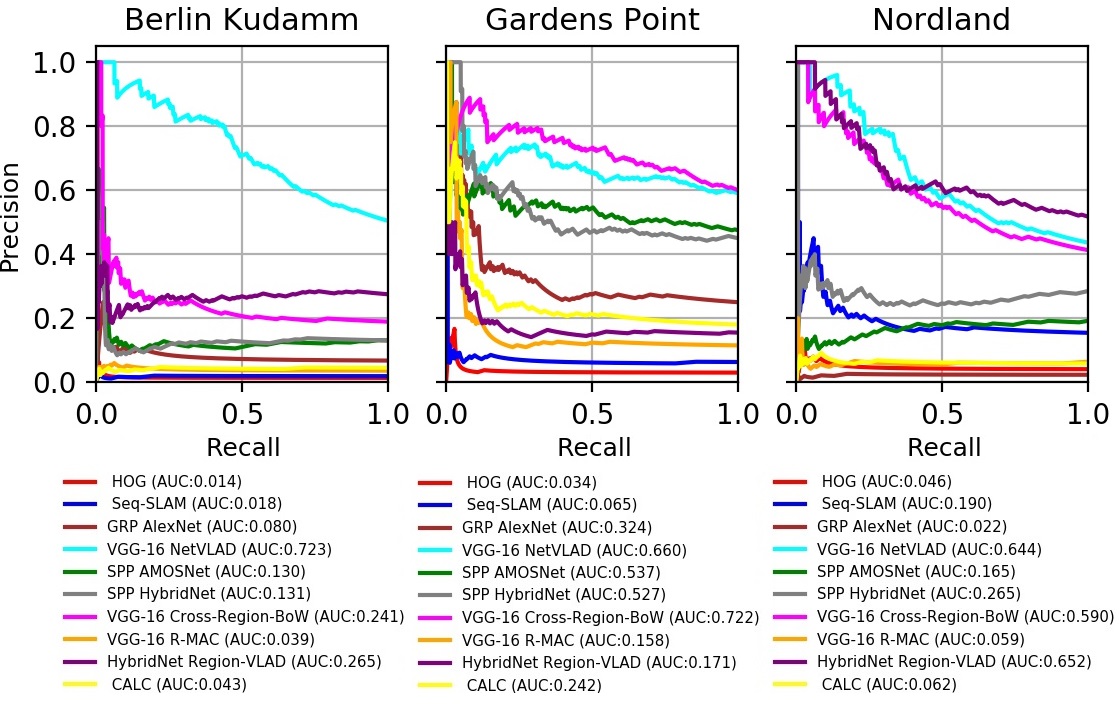}
\end{center}
\caption{AUC under PR curves on the benchmark datasets of all VPR techniques.}
\label{Fig:prCurves}
\end{figure*}

Fig. \ref{Fig:prCurves} shows that NetVLAD achieves state-of-the-art performance on Berlin Kudamm dataset, while Region-VLAD and Cross-Region-BoW follow-up with relatively poor performance. AMOSNet and HybridNet with SPP also achieve nearly similar performance to regions based approaches and suffer due to the extreme viewpoint variation not catered by SPP. It is important to note that due to urban scenario, both the traverses in Berlin Kudamm dataset include dynamic and confusing objects such as vehicles, pedestrians and trees; as illustrated in Fig. \ref{Fig:berlin_kudamm_samples}. These confusing objects and homogeneous scenes lead to perceptual aliasing which coupled with extreme viewpoint variations makes Berlin Kudamm highly challenging for all VPR techniques.

SeqSLAM being velocity dependent has shown inferior results due to the varying speed of camera platform and significant viewpoint variation. One of the reasons for state-of-the-art performance of NetVLAD could be its training on large urban place-centric dataset (Pittsburgh 250K) which exhibits strong lightning and viewpoint variations along with dynamic and confusing objects. This is in contrast to the training datasets of VGG-16 (ImageNet) and HybridNet (SPED). Where, ImageNet is an object detection dataset and is intrinsically not good for place recognition. While SPED does not contain dynamic objects observed in urban road scenes.   

\subsubsection{Gardens Point Dataset}
Although this dataset exhibits strong illumination and viewpoint variations, majority of the VPR approaches perform relatively well. This is due to the distinctive structures captured in both the traverses. Cross-Region-BoW achieves state-of-the-art results while Net-VLAD, HybridNet and AMOSNet also perform nearly well on this dataset.

\subsubsection{Nordland Dataset}
Nordland dataset exhibits strong seasonal variation and synthetic viewpoint change, as illustrated in Fig.\ref{Fig:nordland_samples}. Region-VLAD achieves state-of-the-art performance with Net-VLAD and Cross-Region-BOW also giving comparable results. HybridNet performs better than AMOSNet due to its weights being initialized from the weights of CaffeNet that have been exposed to a variety of scenes available in the ImageNet dataset.

\subsection{Matching Time}
\begin{figure*}[t]
\begin{center}
\includegraphics[width=0.9\linewidth, height=0.4\linewidth]{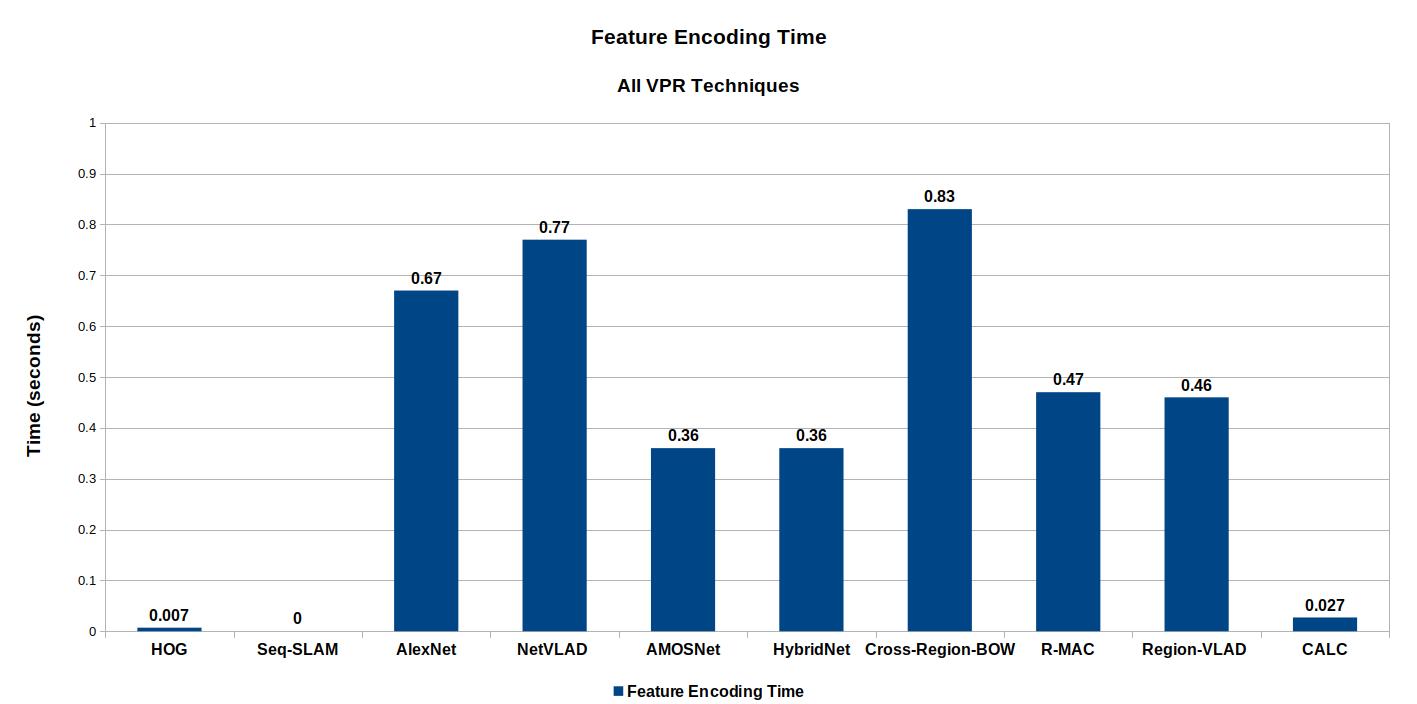}
\end{center}
\caption{Feature encoding time of all VPR techniques are shown in this figure. As expected, neural network based techniques have higher encoding time compared to handcrafted techniques. Although, the matching performance of CALC is lower compared to some of the neural network based VPR techniques, the significantly low encoding time of CALC promises the possibilities of real-time highly accurate VPR in future.}
\label{Fig:feature_encoding_time_all}
\end{figure*}

In real-time VPR systems, matching time is an important factor that needs to be considered when comparing a query image against large number of database images. We show in Fig. \ref{Fig:feature_encoding_time_all}, the feature encoding time for all VPR techniques given a single query image. Seq-SLAM does not extract features from an image but directly uses patch-normalized camera frames for comparison. As expected, CNNs take significantly more time to encode an input image compared to handcrafted feature descriptors. However, convolutional auto-encoder in CALC takes significantly lower time to encode features in comparison to other CNN based VPR techniques. This is because the architecture of CALC is designed specifically for VPR as compared to off-the-shelf CNN architectures employed in other VPR techniques. 

\begin{figure}[t]
\begin{center}
\includegraphics[width=1.0\linewidth]{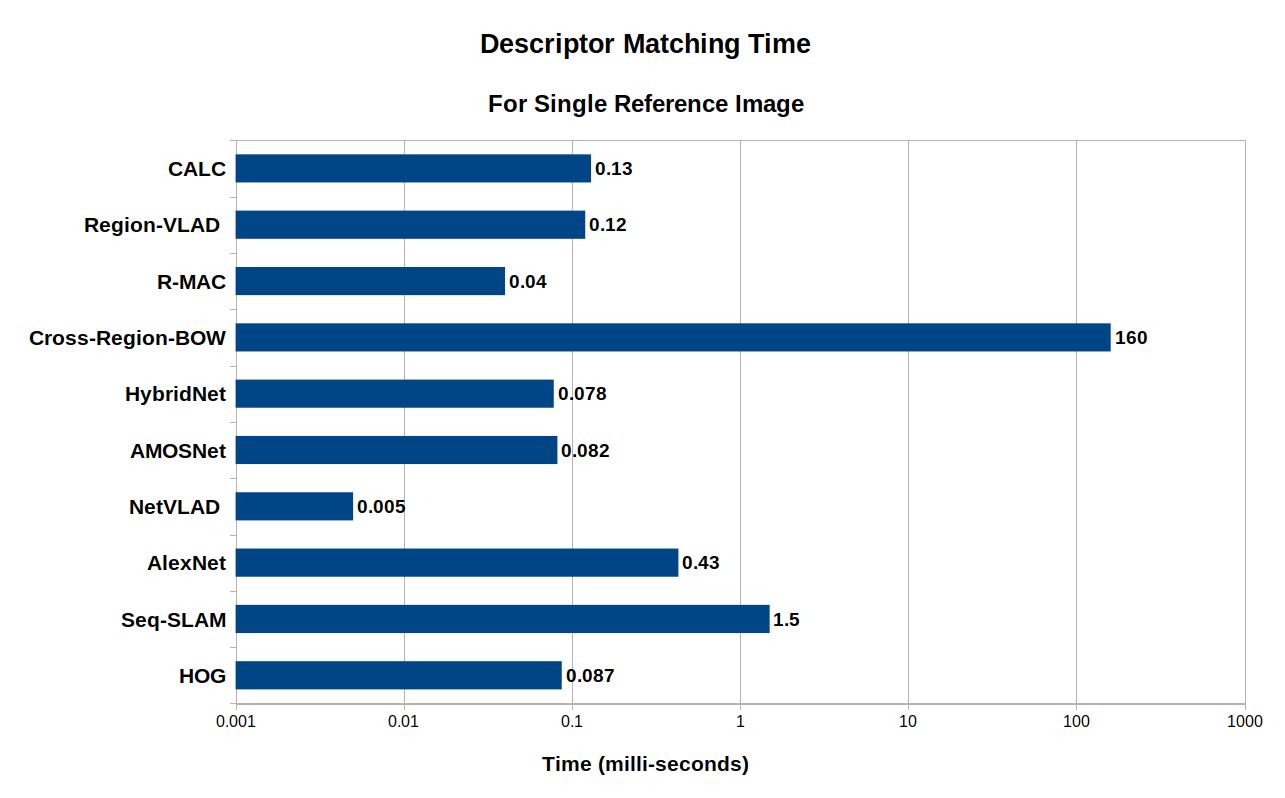}
\end{center}
\caption{Descriptor matching time of all VPR techniques are compared here. Please note that the horizontal axis is in logarithmic scale due to the high variance in-between matching times of different techniques.}
\label{Fig:descriptor_matching_time_all}
\end{figure}

While the feature encoding time is independent of the number of reference images, feature descriptor matching time scales directly with the total number of reference images. Thus, we also show the time taken to match feature descriptors of a query and a reference image in Fig. \ref{Fig:descriptor_matching_time_all}. Please note that Fig. \ref{Fig:descriptor_matching_time_all} uses logarithmic scale on horizontal-axis for clarity. This descriptor matching time can be directly multiplied with the total number of reference images in the database. It is interesting to note that although Cross-Region-BOW achieves good matching performance on different datasets, it suffers from a significantly higher descriptor matching time. This can be associated with the one-to-many nature of Cross-Region-BOW which finds the best matched regions between a query and a reference image.  

\subsection{Memory Footprint}
\begin{figure}[t]
\begin{center}
\includegraphics[width=1.0\linewidth]{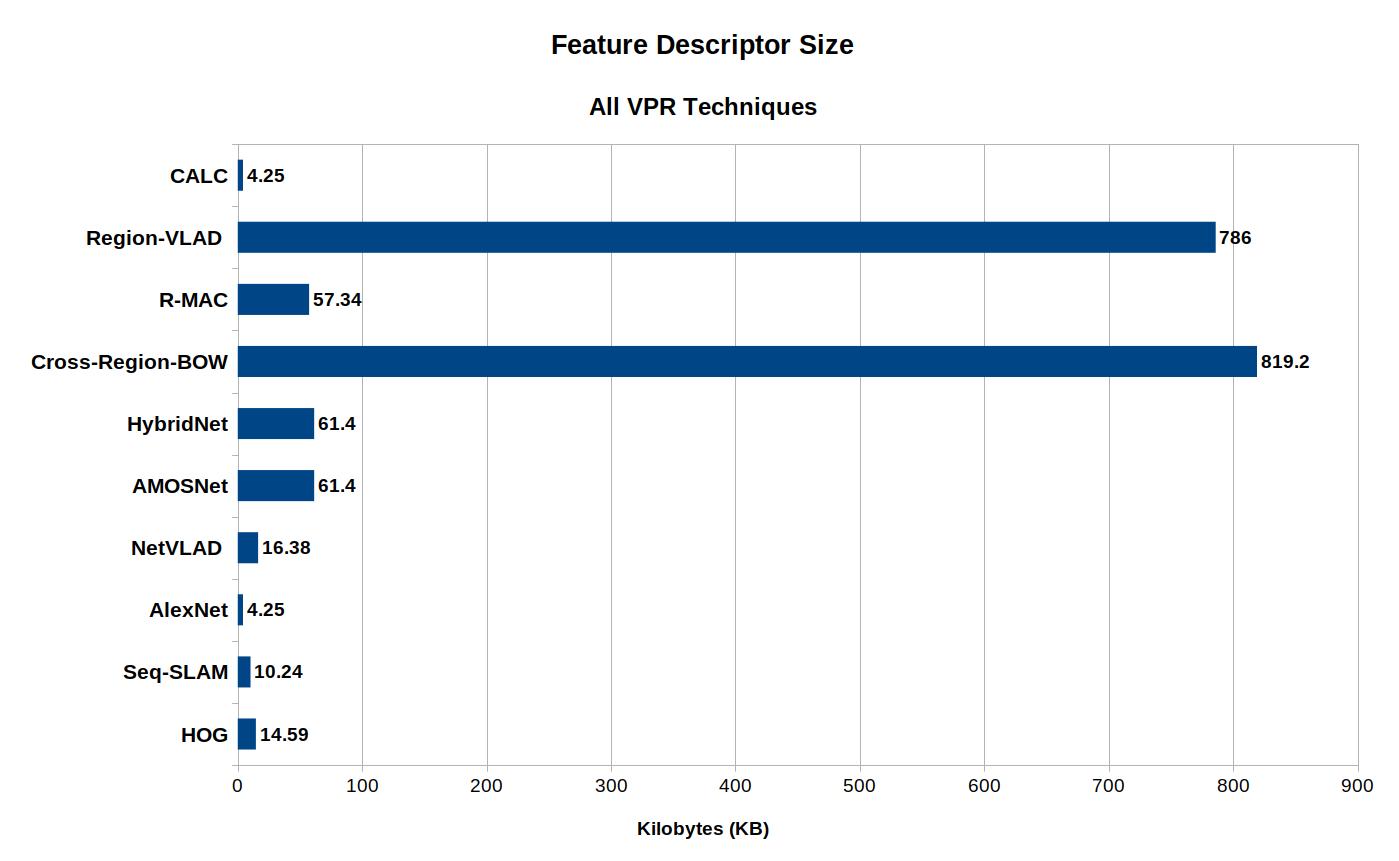}
\end{center}
\caption{Feature descriptor sizes of all VPR techniques are shown here. While this metric has been rarely discussed in VPR literature, it is highly significant for resource-constrained platforms and can hinder the deployment of a VPR technique in field. Thus, highly compact feature descriptors that are encoded in real-time, are condition invariant, repeatable and distinct should be the output of an ideal VPR system.}
\label{Fig:feature_vector_size_all}
\end{figure}

The size of feature descriptors plays an important role when considering the practicality of a VPR technique for deployment in real-world scenarios. Due to limited storage capabilities, compact representations of image descriptors is needed. Thus, while matching performance can be improved by increasing the size of feature descriptor (or number of regions where applicable), it limits the deployment feasibility of such a VPR technique. We have reported the feature vector size of all VPR techniques in Fig. \ref{Fig:feature_vector_size_all}. Cross-Region-BOW and Region-VLAD notably have a large memory footprint compared to other VPR techniques. For Cross-Region-BOW, this can be associated with the number of regions ($200$) that have to be stored, where each region has a descriptor dimension equal to the depth ($512$) of convolutional layer. While in Region-VLAD, the employed VLAD dictionary size is $256$ with each visual-word in the dictionary having a dimension (depth of convolutional layer) of $384$.     

\section{Conclusion}
This paper presented a holistic comparison of $10$ VPR techniques on challenging public datasets. The choice of evaluation datasets, ground truth data, computational platform and comparison metric is kept constant to report the results on a common-ground. In addition to the matching performance and matching time, we report the feature vector size as an important factor for VPR deployment practicality. While neural network based techniques out-perform handcrafted feature descriptors in matching performance, they suffer from higher matching time and larger memory footprint. The performance comparison of neural network based techniques with each other also identifies their lack of generalizability from one evaluation dataset to another. While some VPR techniques can achieve better matching performance in contrast to others, there may be a trade-off between matching performance and computational requirements (i.e. higher matching time and/or memory demand). It is worth noticing that contrary to expectations, increase in VPR performance (for our choice of parameters and datasets) is not observed in a chronological order.

Although our selected evaluation datasets consist of extreme viewpoint, seasonal and illumination variations; they are only moderately sized datasets. The reported results show that VPR techniques are still far from ideal performance even on such medium scale datasets. However, it would be useful to perform a similar evaluation on a large scale dataset which serves as a motivation for future investigation. We hope our work proves as a good reference for VPR research community and fuels incremental performance improvement; thus realizing real-time VPR deployment in autonomous robotics.  

{
\small
\bibliographystyle{ieeetr}
\bibliography{main}

\begin{thebibliography}{10}

\bibitem{cadena2016past}
C.~Cadena and et~al., ``Past, present, and future of simultaneous localization
  and mapping: Toward the robust-perception age,'' {\em IEEE T-RO}, vol.~32,
  no.~6, pp.~1309--1332, 2016.

\bibitem{eade2006scalable}
E.~Eade and T.~Drummond, ``Scalable monocular slam,'' in {\em CVPR},
  pp.~469--476, IEEE Computer Society, 2006.

\bibitem{vprasurvey}
S.~Lowry and et~al., ``Visual place recognition: A survey,'' {\em IEEE T-RO},
  vol.~32, no.~1, pp.~1--19, 2016.

\bibitem{naseer2014robust}
T.~Naseer, L.~Spinello, W.~Burgard, and C.~Stachniss, ``Robust visual robot
  localization across seasons using network flows,'' in {\em Twenty-Eighth AAAI
  Conference on Artificial Intelligence}, 2014.

\bibitem{valgren2010sift}
C.~Valgren and A.~J. Lilienthal, ``Sift, surf \& seasons: Appearance-based
  long-term localization in outdoor environments,'' {\em RAS}, vol.~58, no.~2,
  pp.~149--156, 2010.

\bibitem{pronobis2006discriminative}
A.~Pronobis, B.~Caputo, P.~Jensfelt, and H.~I. Christensen, ``A discriminative
  approach to robust visual place recognition,'' in {\em IROS}, pp.~3829--3836,
  IEEE, 2006.

\bibitem{garg2018lost}
S.~Garg, N.~Suenderhauf, and M.~Milford, ``Lost? appearance-invariant place
  recognition for opposite viewpoints using visual semantics,'' {\em \tt
  arXiv:1804.05526 [cs.RO]}, 2018.

\bibitem{ranganathan2013towards}
A.~Ranganathan, S.~Matsumoto, and D.~Ilstrup, ``Towards illumination invariance
  for visual localization,'' in {\em ICRA}, pp.~3791--3798, IEEE, 2013.

\bibitem{milford2015sequence}
M.~Milford and et~al., ``Sequence searching with deep-learnt depth for
  condition-and viewpoint-invariant route-based place recognition,'' in {\em
  CVPR}, pp.~18--25, 2015.

\bibitem{wang2007simultaneous}
C.-C. Wang and et~al., ``Simultaneous localization, mapping and moving object
  tracking,'' {\em IJRR}, vol.~26, no.~9, pp.~889--916, 2007.

\bibitem{naseer2017semantics}
T.~Naseer, G.~L. Oliveira, T.~Brox, and W.~Burgard, ``Semantics-aware visual
  localization under challenging perceptual conditions,'' in {\em ICRA},
  pp.~2614--2620, IEEE, 2017.

\bibitem{sunderhauf2015performance}
N.~S{\"u}nderhauf and et~al., ``On the performance of convnet features for
  place recognition,'' {\em \tt arXiv:1501.04158 [cs.RO]}, 2015.

\bibitem{sunderhauf2015place}
N.~S{\"u}nderhauf, S.~Shirazi, A.~Jacobson, F.~Dayoub, E.~Pepperell,
  B.~Upcroft, and M.~Milford, ``Place recognition with convnet landmarks:
  Viewpoint-robust, condition-robust, training-free,'' {\em Proceedings of RSS
  XII}, 2015.

\bibitem{panphattarasap2016visual}
P.~Panphattarasap and A.~Calway, ``Visual place recognition using landmark
  distribution descriptors,'' in {\em ACCV}, pp.~487--502, Springer, 2016.

\bibitem{bay2006surf}
H.~Bay, T.~Tuytelaars, and L.~Van~Gool, ``Surf: Speeded up robust features,''
  in {\em ECCV}, pp.~404--417, Springer, 2006.

\bibitem{lowe2004distinctive}
D.~G. Lowe, ``Distinctive image features from scale-invariant keypoints,'' {\em
  IJCV}, vol.~60, no.~2, pp.~91--110, 2004.

\bibitem{stumm2013probabilistic}
E.~Stumm, C.~Mei, and S.~Lacroix, ``Probabilistic place recognition with
  covisibility maps,'' in {\em IROS}, pp.~4158--4163, IEEE, 2013.

\bibitem{murillo2007surf}
A.~C. Murillo, J.~J. Guerrero, and C.~Sagues, ``Surf features for efficient
  robot localization with omnidirectional images,'' in {\em ICRA},
  pp.~3901--3907, IEEE, 2007.

\bibitem{agrawal2008censure}
M.~Agrawal, K.~Konolige, and M.~R. Blas, ``Censure: Center surround extremas
  for realtime feature detection and matching,'' in {\em ECCV}, pp.~102--115,
  Springer, 2008.

\bibitem{rosten2006machine}
E.~Rosten and T.~Drummond, ``Machine learning for high-speed corner
  detection,'' in {\em ECCV}, pp.~430--443, Springer, 2006.

\bibitem{sivic2003video}
J.~Sivic and A.~Zisserman, ``Video google: A text retrieval approach to object
  matching in videos,'' in {\em null}, p.~1470, IEEE, 2003.

\bibitem{oliva2006building}
A.~Oliva and A.~Torralba, ``Building the gist of a scene: The role of global
  image features in recognition,'' {\em Progress in brain research}, vol.~155,
  pp.~23--36, 2006.

\bibitem{murillo2009experiments}
A.~C. Murillo and J.~Kosecka, ``Experiments in place recognition using gist
  panoramas,'' in {\em ICCV Workshops}, pp.~2196--2203, IEEE, 2009.

\bibitem{singh2010visual}
G.~Singh and J.~Kosecka, ``Visual loop closing using gist descriptors in
  manhattan world,'' in {\em ICRA Omnidirectional Vision Workshop}, 2010.

\bibitem{badino2012real}
H.~Badino, D.~Huber, and T.~Kanade, ``Real-time topometric localization,'' in
  {\em ICRA}, pp.~1635--1642, IEEE, 2012.

\bibitem{MERL_TR9403}
W.~T. Freeman and M.~Roth, ``Orientation histograms for hand gesture
  recognition,'' Tech. Rep. TR94-03, MERL - Mitsubishi Electric Research
  Laboratories, Cambridge, MA 02139, Dec. 1994.

\bibitem{dalal2005histograms}
N.~Dalal and B.~Triggs, ``Histograms of oriented gradients for human
  detection,'' in {\em CVPR}, vol.~1, pp.~886--893, IEEE Computer Society,
  2005.

\bibitem{mcmanus2014scene}
C.~McManus, B.~Upcroft, and P.~Newmann, ``Scene signatures: Localised and
  point-less features for localisation,'' 2014.

\bibitem{seqslam}
M.~J. Milford and G.~F. Wyeth, ``Seqslam: Visual route-based navigation for
  sunny summer days and stormy winter nights,'' in {\em ICRA}, pp.~1643--1649,
  IEEE, 2012.

\bibitem{tolias2015particular}
G.~Tolias, R.~Sicre, and H.~J{\'e}gou, ``Particular object retrieval with
  integral max-pooling of cnn activations,'' {\em \tt arXiv:1511.05879
  [cs.CV]}, 2015.

\bibitem{liu2017cross}
L.~Liu, C.~Shen, and A.~van~den Hengel, ``Cross-convolutional-layer pooling for
  image recognition,'' {\em IEEE T-PAMI}, vol.~39, no.~11, pp.~2305--2313,
  2017.

\bibitem{chen2014convolutional}
Z.~Chen, O.~Lam, A.~Jacobson, and M.~Milford, ``Convolutional neural
  network-based place recognition,'' {\em \tt arXiv:1411.1509 [cs.CV]}, 2014.

\bibitem{sermanet2013overfeat}
P.~Sermanet, D.~Eigen, X.~Zhang, M.~Mathieu, R.~Fergus, and Y.~LeCun,
  ``Overfeat: Integrated recognition, localization and detection using
  convolutional networks,'' {\em \tt arXiv:1312.6229 [cs.CV]}, 2013.

\bibitem{chen2017deep}
Z.~Chen and et~al., ``Deep learning features at scale for visual place
  recognition,'' in {\em ICRA}, pp.~3223--3230, IEEE, 2017.

\bibitem{krizhevskyimagenet}
A.~Krizhevsky, I.~Sutskever, and G.~Hinton, ``Imagenet classification with deep
  convolutional networks,'' in {\em NIPS}, pp.~1097--1105.

\bibitem{deng2009imagenet}
J.~Deng, W.~Dong, R.~Socher, L.-J. Li, K.~Li, and L.~Fei-Fei, ``Imagenet: A
  large-scale hierarchical image database,'' in {\em CVPR}, pp.~248--255, IEEE,
  2009.

\bibitem{babenko2015aggregating}
A.~Babenko and V.~Lempitsky, ``Aggregating local deep features for image
  retrieval,'' in {\em ICCV}, pp.~1269--1277, 2015.

\bibitem{jaderberg2015spatial}
M.~Jaderberg, K.~Simonyan, A.~Zisserman, {\em et~al.}, ``Spatial transformer
  networks,'' in {\em Advances in NIPS}, pp.~2017--2025, 2015.

\bibitem{arandjelovic2016netvlad}
R.~Arandjelovic and et~al, ``Netvlad: Cnn architecture for weakly supervised
  place recognition,'' in {\em CVPR}, pp.~5297--5307, 2016.

\bibitem{merrill2018lightweight}
N.~Merrill and G.~Huang, ``Lightweight unsupervised deep loop closure,'' {\em
  \tt arXiv:1805.07703 [cs.RO]}, 2018.

\bibitem{torii2013visual}
A.~Torii, J.~Sivic, T.~Pajdla, and M.~Okutomi, ``Visual place recognition with
  repetitive structures,'' in {\em CVPR}, pp.~883--890, 2013.

\bibitem{torii201524}
A.~Torii, R.~Arandjelovic, J.~Sivic, M.~Okutomi, and T.~Pajdla, ``24/7 place
  recognition by view synthesis,'' in {\em CVPR}, pp.~1808--1817, 2015.

\bibitem{chen2018learning}
Z.~Chen, L.~Liu, I.~Sa, Z.~Ge, and M.~Chli, ``Learning context flexible
  attention model for long-term visual place recognition,'' {\em IEEE RA-L},
  vol.~3, no.~4, pp.~4015--4022, 2018.

\bibitem{khaliqholistic}
A.~Khaliq, S.~Ehsan, M.~Milford, and K.~McDonald~Maier, ``A holistic visual
  place recognition approach using lightweight cnns for severe viewpoint and
  appearance changes,'' {\em \tt arXiv:1811.03032 [cs.RO]}, 2018.

\bibitem{facil2019condition}
J.~M. Facil, D.~Olid, L.~Montesano, and J.~Civera, ``Condition-invariant
  multi-view place recognition,'' {\em \tt arXiv:1902.09516 [cs.CV]}, 2019.

\bibitem{hausler2019multi}
S.~Hausler, A.~Jacobson, and M.~J. Milford, ``Multi-process fusion: Visual
  place recognition using multiple image processing methods,'' {\em IEEE RA-L},
  2019.

\bibitem{chen2017only}
Z.~Chen, F.~Maffra, I.~Sa, and M.~Chli, ``Only look once, mining distinctive
  landmarks from convnet for visual place recognition,'' in {\em IROS},
  pp.~9--16, IEEE, 2017.

\bibitem{simonyan2014very}
K.~Simonyan and A.~Zisserman, ``Very deep convolutional networks for
  large-scale image recognition,'' {\em \tt arXiv:1409.1556 [cs.CV]}, 2014.

\bibitem{khaliq2018holistic}
A.~Khaliq, S.~Ehsan, M.~Milford, and K.~McDonald-Maier, ``A holistic visual
  place recognition approach using lightweight cnns for severe viewpoint and
  appearance changes,'' {\em \tt arXiv:1811.03032 [cs.RO]}, 2018.

\bibitem{cieslewski2018data}
T.~Cieslewski, S.~Choudhary, and D.~Scaramuzza, ``Data-efficient decentralized
  visual slam,'' in {\em ICRA}, pp.~2466--2473, IEEE, 2018.

\bibitem{chen2017onlyCode}
``Region-bow.'' \url{https://github.com/scutzetao/IROS2017\_OnlyLookOnce}.

\bibitem{tolias2015particularRMACCode}
``Rmac.'' \url{https://github.com/gtolias/rmac}.

\bibitem{3}
M.~Warren, D.~McKinnon, H.~He, and B.~Upcroft, ``Unaided stereo vision based
  pose estimation,'' in {\em ACRA}, 2010.

\bibitem{4}
A.~J. Glover, W.~P. Maddern, M.~J. Milford, and G.~F. Wyeth, ``Fab-map+
  ratslam: Appearance-based slam for multiple times of day,'' in {\em ICRA},
  pp.~3507--3512, IEEE, 2010.

\bibitem{5}
M.~Milford and et~al., ``Sequence searching with deep-learnt depth for
  condition-and viewpoint-invariant route-based place recognition,'' in {\em
  CVPR Workshops}, pp.~18--25, 2015.

\bibitem{nordlanddataset}
S.~Skrede, ``Nordland dataset.'' \url{https://bit.ly/2QVBOym}, 2013.

\bibitem{8}
A.~Geiger, P.~Lenz, C.~Stiller, and R.~Urtasun, ``Vision meets robotics: The
  kitti dataset,'' {\em IJRR}, 2013.

\bibitem{10}
T.~Naseer, W.~Burgard, and C.~Stachniss, ``Robust visual localization across
  seasons,'' {\em IEEE T-RO}, vol.~34, no.~2, pp.~289--302, 2018.

\end{thebibliography}
}

\end{document}